\documentclass{article}

\usepackage{microtype}
\usepackage{graphicx}
\usepackage{subcaption}
\usepackage{booktabs}
\usepackage{hyperref}
\usepackage{placeins}

\usepackage[preprint]{icml2026}

\makeatletter
\renewcommand{\ICML@preprint}{\textit{Preprint.}}
\makeatother

\usepackage{amsmath}
\usepackage{amssymb}
\usepackage{mathtools}
\usepackage{amsthm}
\usepackage[capitalize,noabbrev]{cleveref}

\hypersetup{
  pdftitle={Giga-Embeddings: Mixture-of-Experts Encoders for High-Throughput Text Embeddings},
  pdfauthor={Egor Kolodin, Egor Krasnoperov, Evgeniy Kosarev, Fyodor Minkin},
  pdfsubject={Preprint}
}

\icmltitlerunning{Giga-Embeddings: High-Throughput MoE Text Encoders}

\begin{document}

\twocolumn[
  \icmltitle{Giga-Embeddings: Mixture-of-Experts Encoders for\texorpdfstring{\\}{ }High-Throughput Text Embeddings}

  \begin{icmlauthorlist}
    \icmlauthor{Egor Kolodin}{aff1,aff2}
    \icmlauthor{Egor Krasnoperov}{aff2}
    \icmlauthor{Evgeniy Kosarev}{aff2,aff3}
    \icmlauthor{Fyodor Minkin}{aff1,aff2}
  \end{icmlauthorlist}

  \icmlaffiliation{aff1}{MIPT}
  \icmlaffiliation{aff2}{SaluteDevices}
  \icmlaffiliation{aff3}{MSU}
  \icmlcorrespondingauthor{Egor Kolodin}{kolodin.ei@phystech.edu}
  \icmlkeywords{text embeddings, mixture of experts, sparse activation, inference throughput, knowledge distillation}

  \vskip 0.3in
]

\printAffiliationsAndNotice{}

\begin{abstract}
We introduce Giga-Embeddings, a family of text embedding models designed to combine strong retrieval quality with efficient serving.
Its largest member is a sparse 10B-parameter Mixture-of-Experts encoder that preserves substantial model capacity while activating only a fraction of it for each token.
Across English, Russian, multilingual, and code MTEB benchmarks, the MoE model achieves the strongest aggregate performance within the family on all four evaluated suites.
At an input length of 1024 tokens under our benchmark environment, its throughput is 25\% higher than that of the dense 3B model and 1.56--2.65$\times$ that of the evaluated external systems.
The family further includes a dense 3B model and a distilled 480M model, extending this quality--efficiency trade-off to tighter compute and memory budgets.
\end{abstract}

\section{Introduction}
\label{sec:introduction}

Text embeddings support semantic search, retrieval-augmented generation, clustering, classification, and semantic textual similarity.
MTEB~\citep{muennighoff2023mteb} and its multilingual expansion MMTEB~\citep{enevoldsen2025mmteb} provide broad evaluation suites for these use cases.
Recent embedding systems adapt decoder-only language models into bidirectional encoders and train them with large-scale contrastive objectives~\citep{behnamghader2024llm2vec,wang2024improving,lee2025nvembed,kolodin2025gigaembeddings}.
Although scaling the backbone generally improves representation quality, billion-parameter encoders are costly to store and serve.

Knowledge distillation offers one route to smaller encoders.
For embedding models, supervision can be transferred by matching embeddings or by matching the teacher's relative similarity scores over candidate sets~\citep{reimers2020multilingual,xu2023distillcse,zhao2026kalm}.
The latter is attractive when the teacher and student have different hidden dimensions or architectures.
It is also architecture-agnostic: a dense or sparse Mixture-of-Experts (MoE) teacher can supervise the same listwise objective through its final similarity scores.

We present two complementary approaches to efficient text embeddings: sparse activation for high-capacity models and similarity-distribution distillation for compact encoders.
The 10B MoE model activates approximately 1.8B parameters per token and achieves the highest measured throughput among the compared systems.
The distilled 480M model provides a compact alternative while remaining within 2.41--6.92 points of the dense 3B model across the four evaluation suites.
Our contributions are:
\begin{enumerate}
  \item Giga-Embeddings-10B-A1.8B, a sparse encoder that achieves the best aggregate scores in the model family, leads the external Russian comparison, and delivers the highest measured throughput at all three sequence lengths;
  \item a dimension-agnostic similarity-distribution distillation method that produces a 480M encoder scoring 70.98 on Russian MTEB, outperforming the 70.95 score of FRIDA while using 42\% fewer parameters (480M vs.\ 823M);
  \item the open-source release of all three model checkpoints: \href{https://huggingface.co/ai-sage/Giga-Embeddings-instruct-480M-0826}{480M}, \href{https://huggingface.co/ai-sage/Giga-Embeddings-instruct-3B-0826}{3B}, and \href{https://huggingface.co/ai-sage/Giga-Embeddings-instruct-10B-A1.8B-0826}{10B-A1.8B}.
\end{enumerate}

\section{Related Work}
\label{sec:related}

\paragraph{General-purpose text embeddings.}
Sentence-BERT~\citep{reimers2019sentence} and SimCSE~\citep{gao2021simcse} established efficient bi-encoder and contrastive recipes.
E5~\citep{wang2022text}, GTE~\citep{li2023towards}, BGE~\citep{xiao2023cpack}, and Nomic Embed~\citep{nussbaum2024nomic} scaled data and negative mining.
Instruction-conditioned systems such as INSTRUCTOR~\citep{su2023instructor}, E5-Mistral~\citep{wang2024improving}, GritLM~\citep{muennighoff2025gritlm}, and NV-Embed~\citep{lee2025nvembed} adapt large language models to heterogeneous embedding tasks.
Recent multilingual LLM-based embedders include Qwen3-Embedding~\citep{zhang2025qwen3embedding}, which uses the same backbone family as our dense encoders, and Gemini Embedding~\citep{lee2025gemini}.

\paragraph{Embedding distillation.}
Classical knowledge distillation matches softened output distributions~\citep{hinton2015distilling}, and compact encoders have long been obtained by direct compression of BERT-style models~\citep{sun2019patient,sanh2019distilbert,jiao2020tinybert,sun2020mobilebert}.
In dense retrieval, cross-architecture and listwise similarity distillation are well established: Margin-MSE transfers cross-encoder margins to bi-encoders~\citep{hofstaetter2020improving}, RocketQAv2 applies dynamic listwise distillation between retriever and re-ranker~\citep{ren2021rocketqav2}, and Gecko distills knowledge from large language models into a compact embedder~\citep{lee2024gecko}.
For sentence embeddings, \citet{reimers2020multilingual} transfer representations from a monolingual teacher to a multilingual student, while DistillCSE~\citep{xu2023distillcse} studies contrastive self-distillation and teacher-logit variance.
GUIDE~\citep{trinh2025guide} combines guided initialization with representation transfer.
Recent contrastive-plus-distillation recipes for compact embedders include KaLM-Embedding-V2~\citep{zhao2026kalm}, which weights contrastive loss against a KL divergence between teacher and student similarity distributions, and jina-embeddings-v5-text~\citep{akram2026jina}, which combines distillation with task-targeted contrastive training.
Our method similarly combines contrastive learning with KL divergence between teacher and student similarity distributions.

\paragraph{Sparse teachers and dense students.}
MoE models increase total capacity while activating a subset of experts per token~\citep{shazeer2017outrageously,fedus2022switch,jiang2024mixtral}.
Sparse-to-dense transfer has been studied for generative models; for example, \citet{xue2022one} compare several ways of gathering expert weights before distillation, and \citet{fedus2022switch} describe sparse-to-dense distillation with partial retention of the sparse quality gain.
Because final-output distillation operates on similarity scores, it can transfer knowledge from an MoE teacher to a dense student without matching internal routing decisions or expert states.

\section{Model and Training Recipe}
\label{sec:method}

\begin{table}[!t]
  \caption{Stage-specific training settings. Each fine-tuning and multitask sample contains one positive and seven negatives, giving 8,192 candidate texts per global batch.}
  \label{tab:training}
  \centering
  \footnotesize
  \setlength{\tabcolsep}{3pt}
  \begin{tabular}{lccc}
    \toprule
    \textbf{Setting} & \textbf{Pretrain} & \textbf{Finetune} & \textbf{Multitask} \\
    \midrule
    Global batch size       & 16,384 & 1,024 & 1,024 \\
    Candidates/sample       & in-batch & 1 pos. + 7 neg. & 1 pos. + 7 neg. \\
    In-batch negatives      & yes & no & no \\
    \bottomrule
  \end{tabular}
\end{table}

\subsection{Encoder adaptation}

Following the earlier Giga-Embeddings recipe~\citep{kolodin2025gigaembeddings}, the model family adapts decoder-only language models for embedding generation.
The causal attention mask is replaced by a fully visible bidirectional mask, as in LLM2Vec~\citep{behnamghader2024llm2vec}, so that each token can use both left and right context.
We use mean pooling to aggregate token states into a fixed-dimensional vector and normalize the resulting vector before cosine-similarity scoring.
The instruction templates follow the earlier Giga-Embeddings recipe~\citep{kolodin2025gigaembeddings} and distinguish asymmetric retrieval inputs from symmetric semantic-similarity inputs.
The \href{https://huggingface.co/ai-sage/Giga-Embeddings-instruct-480M-0826}{480M}, \href{https://huggingface.co/ai-sage/Giga-Embeddings-instruct-3B-0826}{3B}, and \href{https://huggingface.co/ai-sage/Giga-Embeddings-instruct-10B-A1.8B-0826}{10B-A1.8B} checkpoints and instruction set are publicly available.

The 480M, 3B, and 10B-A1.8B encoders produce embeddings with dimensions 1024, 2048, and 1536, respectively.
The 480M and 3B models are dense bidirectional Qwen3 encoders~\citep{zhang2025qwen3embedding}, while the largest model is a bidirectional DeepSeekMoE-style encoder~\citep{dai2024deepseekmoe} with 10B total parameters, 64 routed experts, one shared expert, and top-4 routing.
Its release designation, A1.8B, denotes approximately 1.8B active parameters per token.
We use ``480M'', ``3B'', and ``10B-A1.8B'' as short model-family names.

\subsection{Three-stage contrastive training}

As in the first Giga-Embeddings release~\citep{kolodin2025gigaembeddings}, training has three stages: broad contrastive pre-training, retrieval fine-tuning with hard negatives, and multitask fine-tuning for retrieval, classification, clustering, and semantic similarity.
The training mixture combines open-source datasets used in the earlier Giga-Embeddings recipe~\citep{kolodin2025gigaembeddings} with non-public datasets that cannot be disclosed under their contractual terms.
Across all stages, the learning rate starts at $3\times10^{-5}$ and follows a cosine decay schedule with a 1\% warmup fraction and a minimum learning rate of $10^{-7}$.
We use weight decay of 0.01, Adam coefficients $\beta_1=0.9$ and $\beta_2=0.999$, and an InfoNCE temperature of 0.02.
Fine-tuning runs for one epoch and multitask training for three epochs.
The stage-specific settings are summarized in \Cref{tab:training}.

Our contrastive-loss formulation follows Qwen3-Embedding~\citep{zhang2025qwen3embedding}.
Pre-training uses in-batch InfoNCE, includes query--query and document--document contrastive terms, and masks detected false negatives.
Fine-tuning and multitask training disable in-batch negatives and use the explicit eight-candidate groups described in \Cref{tab:training}.
For a batch of $B$ queries, let $\mathcal{C}_i$ be the fixed candidate set for query $q_i$, let $p(i)$ index its positive candidate, and let $s_{ij}$ be the student cosine similarity.

\begin{figure*}[!t]
  \centering
  \includegraphics[width=\textwidth]{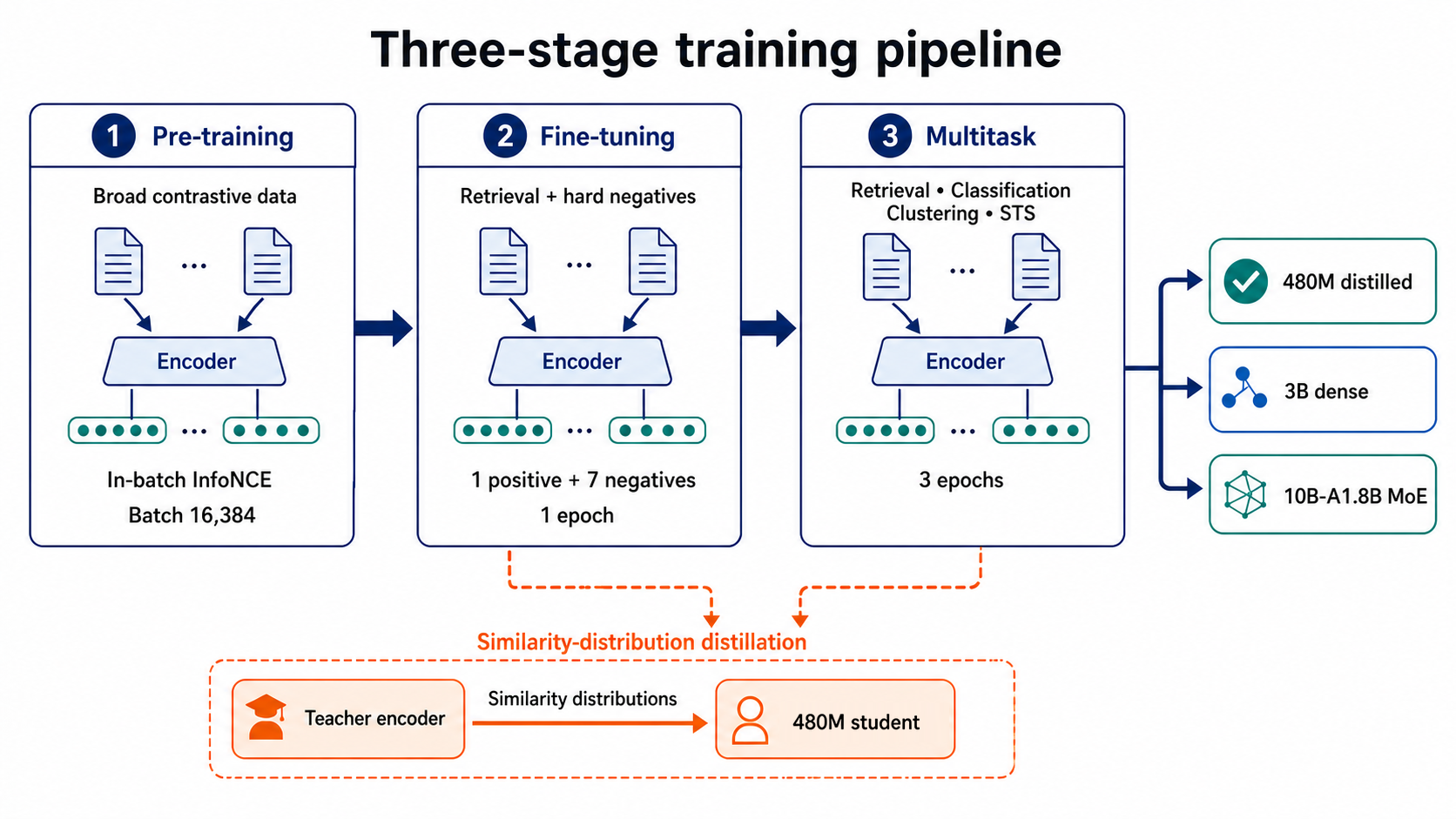}
  \caption{Three-stage embedding training pipeline. All models undergo contrastive pre-training, retrieval fine-tuning, and multitask fine-tuning. Similarity-distribution distillation is applied only to the 480M student during the fine-tuning and multitask stages.}
  \label{fig:training-pipeline}
\end{figure*}

\begin{samepage}
We use batch-mean InfoNCE~\citep{oord2018representation}:
\begin{equation}
  \mathcal{L}_{\mathrm{NCE}} = -\frac{1}{B}\sum_{i=1}^{B}
  \log \frac{\exp(s_{i,p(i)}/\tau)}
  {\sum_{j \in \mathcal{C}_i}\exp(s_{ij}/\tau)}.
  \label{eq:nce}
\end{equation}
\end{samepage}
Hard negatives for retrieval fine-tuning are mined from a predefined retrieval-rank range.

\subsection{Similarity-distribution distillation}
\label{sec:distill}

We apply similarity-distribution distillation exclusively to the 480M model during retrieval fine-tuning and multitask fine-tuning.
Its pre-training stage uses only the contrastive objective, and the 3B and 10B-A1.8B models are trained without distillation.
For both distilled stages, $\lambda=0.3$, $\tau_{\mathrm{KD}}=0.05$, and the contrastive temperature is 0.02.
Because the objective aligns distributions over query--candidate similarity scores rather than embedding coordinates, the teacher and student may use different embedding dimensions.
By contrast, direct MSE or cosine-similarity alignment between teacher and student embedding vectors requires equal dimensions or an additional learned projection.
Let $t_{ij}$ be the teacher cosine similarity for the same support $\mathcal{C}_i$.
Teacher and student distributions are
\begin{align}
P_T(j\mid q_i) &= \frac{\exp(t_{ij}/\tau_{\mathrm{KD}})}
 {\sum_{k\in\mathcal{C}_i}\exp(t_{ik}/\tau_{\mathrm{KD}})}, \\
P_S(j\mid q_i) &= \frac{\exp(s_{ij}/\tau_{\mathrm{KD}})}
 {\sum_{k\in\mathcal{C}_i}\exp(s_{ik}/\tau_{\mathrm{KD}})}.
\end{align}
The distillation term is a batch mean over identical supports:
\begin{equation}
  \mathcal{L}_{\mathrm{KD}} = \frac{1}{B}\sum_{i=1}^{B}
  D_{\mathrm{KL}}\!\left(P_T(\cdot\mid q_i)\,\|\,P_S(\cdot\mid q_i)\right).
  \label{eq:kd}
\end{equation}
The combined objective is
\begin{equation}
  \mathcal{L}=\lambda\mathcal{L}_{\mathrm{KD}}
  +(1-\lambda)\mathcal{L}_{\mathrm{NCE}}.
  \label{eq:combined}
\end{equation}
The KL term transfers the teacher's relative similarity structure over each candidate set, while InfoNCE preserves direct contrastive supervision.
Because the loss depends only on final similarity scores, it does not require access to the teacher's internal architecture or routing decisions.

% Evaluation tables are declared here so they appear near the section without
% interrupting the text flow below the single-column training table.
\begin{table*}[!t]
  \centering
  \begin{minipage}[t]{0.35\textwidth}
    \vspace{0pt}
    \captionof{table}{Model architectures. ``Active'' denotes parameters used per token.}
    \label{tab:architecture}
    \centering
    \small
    \setlength{\tabcolsep}{3.5pt}
    \begin{tabular}{lrrr}
      \toprule
      \textbf{Type} & \textbf{Total} & \textbf{Active} & \textbf{Dim.} \\
      \midrule
      Dense      & 480M & 480M & 1024 \\
      Dense      & 3B & 3B & 2048 \\
      Mixture of Experts & 10B & $\sim$1.8B & 1536 \\
      \bottomrule
    \end{tabular}
  \end{minipage}
  \hfill
  \begin{minipage}[t]{0.62\textwidth}
    \vspace{0pt}
    \captionof{table}{Task-macro MTEB scores (\%) over 41 English, 23 Russian, 131 multilingual, and 12 code tasks. Bold marks the best result within the model family.}
    \label{tab:main}
    \centering
    \small
    \setlength{\tabcolsep}{3.5pt}
    \begin{tabular}{rrrrrr}
      \toprule
      \textbf{Scale} & \textbf{Dim.} & \textbf{English} & \textbf{Russian} & \textbf{Multilingual} & \textbf{Code} \\
      \midrule
      480M       & 1024 & 69.52 & 70.98 & 56.97 & 72.87 \\
      3B         & 2048 & 71.93 & 74.56 & 63.89 & 76.93 \\
      10B-A1.8B  & 1536 & \textbf{72.23} & \textbf{74.98} & \textbf{65.64} & \textbf{78.41} \\
      \bottomrule
    \end{tabular}
  \end{minipage}
\end{table*}

\begin{table*}[!t]
  \caption{Token throughput (thousands of tokens per second) with vLLM at fixed input lengths. Relative throughput is measured at 1024 tokens and normalized to our 10B-A1.8B-0826 model.}
  \label{tab:throughput}
  \centering
  \small
  \setlength{\tabcolsep}{5pt}
  \begin{tabular}{p{8.0cm}rrrr}
    \toprule
    \textbf{Model (vLLM)} & \textbf{512 tokens} & \textbf{1024 tokens} & \textbf{2048 tokens} & \textbf{Relative} \\
    \midrule
    Qwen3 Embedding 4B
      & 70.1k & 73.2k & 71.2k & 0.64$\times$ \\
    F2LLM-v2-8B
      & 43.2k & 43.4k & 42.6k & 0.38$\times$ \\
    Nemotron 8B
      & 42.6k & 43.2k & 41.7k & 0.38$\times$ \\
    \midrule
    \textbf{Ours: 3B-0826}
      & 87.9k & 91.5k & 90.4k & 0.80$\times$ \\
    \textbf{Ours: 10B-A1.8B-0826}
      & \textbf{112.6k} & \textbf{114.5k} & \textbf{102.3k} & \textbf{1.00$\times$} \\
    \bottomrule
  \end{tabular}
\end{table*}

\section{Evaluation}
\label{sec:evaluation}

\subsection{Evaluation protocol}

We evaluate each benchmark using the tasks, subsets, and evaluation splits defined by MTEB.
Scores are first aggregated within each task and then averaged uniformly across tasks to obtain the suite-level result.
This prevents tasks with more languages, subsets, or splits from receiving additional weight.
The Russian suite follows ruMTEB~\citep{snegirev2025rumteb}, while the multilingual suite follows MMTEB~\citep{enevoldsen2025mmteb}.
We use the same aggregation procedure for all models.

Assuming fp16 weights, the 480M, 3B, and 10B-A1.8B models require approximately 0.96, 6, and 20~GB of weight storage, respectively.
The distilled model therefore uses about 4.8\% as much weight storage as the MoE model.
These theoretical estimates exclude activations and other runtime state.

\subsection{Aggregate results}

\Cref{tab:main} presents task-macro scores for all three models.
The 480M distilled model reaches 69.52 on English, 70.98 on Russian, 56.97 on Multilingual, and 72.87 on Code.
Relative to the 3B model, its gaps are 2.41, 3.58, 6.92, and 4.06 points, respectively, with the largest compression gap on the multilingual suite.

The 10B-A1.8B model achieves the highest score within the family on every benchmark.
The improvement over 3B is modest on English (+0.30) and Russian (+0.42), and larger on Multilingual (+1.75) and Code (+1.48).
Because each model is evaluated once, the sub-one-point differences on English and Russian should be interpreted cautiously.

\subsection{Distillation ablation}

\begin{table}[!t]
  \caption{Distillation ablation for the 480M model on the English, Russian, and Code MTEB suites (\%).}
  \label{tab:distillation-ablation}
  \centering
  \small
  \setlength{\tabcolsep}{3.5pt}
  \begin{tabular}{lrrr}
    \toprule
    \textbf{Model} & \textbf{English} & \textbf{Russian} & \textbf{Code} \\
    \midrule
    480M, no distillation & 69.43 & 70.86 & 72.65 \\
    480M, distilled       & \textbf{69.52} & \textbf{70.98} & \textbf{72.87} \\
    \bottomrule
  \end{tabular}
\end{table}

The matched ablation in \Cref{tab:distillation-ablation} isolates the contribution of teacher supervision to the 480M model.
Similarity-distribution distillation improves English, Russian, and Code scores by 0.09, 0.12, and 0.22 points, respectively.
The gains are consistent across all three suites, with the largest improvement on Code.

\subsection{Inference throughput}

The MoE serving configuration has the highest measured throughput at every evaluated length.
Relative to the dense Giga-Embeddings 3B model, its throughput is 1.28$\times$, 1.25$\times$, and 1.13$\times$ as high at 512, 1024, and 2048 tokens, respectively.
All models in \Cref{tab:throughput} use vLLM; the measurements remain specific to the benchmark environment and do not isolate the contribution of sparse activation.
The 480M model was not included in the throughput measurements, so we do not compare its inference speed.
For this model, the only efficiency-related quantity reported is its estimated fp16 weight storage (0.96~GB), which measures model size rather than runtime efficiency.

\subsection{External baseline comparison}

\begin{table*}[!t]
  \caption{Comparison with compact models on MTEB benchmark suites (\%). Public baselines use official MTEB Leaderboard \texttt{meanTask} values retrieved on 2026-08-08; our result is from \Cref{tab:main}. Bold marks the best score in each column, and a dash denotes an unavailable score.}
  \label{tab:baselines}
  \centering
  \small
  \setlength{\tabcolsep}{5pt}
  \begin{tabular}{lrrcccc}
    \toprule
    \textbf{Model} & \textbf{Params} & \textbf{Dim.} & \textbf{Russian} & \textbf{English} & \textbf{Multilingual} & \textbf{Code} \\
    \midrule
    sergeyzh/BERTA                              & 128M & 768  & 69.39 & --- & --- & --- \\
    google/embeddinggemma-300m                  & 308M & 768  & 65.19 & 69.67 & 61.15 & 68.76 \\
    codefuse-ai/F2LLM-v2-0.6B                   & 596M & 1024 & 65.97 & 69.97 & 62.74 & \textbf{77.41} \\
    Qwen/Qwen3-Embedding-0.6B                   & 596M & 1024 & 64.29 & \textbf{70.47} & \textbf{64.34} & 75.42 \\
    ai-forever/FRIDA                            & 823M & 1536 & 70.95 & --- & --- & --- \\
    \midrule
    \textbf{Ours: 480M-0826}                    & 480M & 1024 & \textbf{70.98} & 69.52 & 56.97 & 72.87 \\
    \bottomrule
  \end{tabular}
\end{table*}

\begin{table*}[!t]
  \caption{Comparison with larger models on MTEB benchmark suites (\%). Public baselines use official MTEB Leaderboard \texttt{meanTask} values retrieved on 2026-08-08; our results are from \Cref{tab:main}. Bold marks the best score in each column.}
  \label{tab:large-baselines}
  \centering
  \small
  \setlength{\tabcolsep}{4.5pt}
  \begin{tabular}{lrrcccc}
    \toprule
    \textbf{Model} & \textbf{Params} & \textbf{Dim.} & \textbf{Russian} & \textbf{English} & \textbf{Multilingual} & \textbf{Code} \\
    \midrule
    codefuse-ai/F2LLM-v2-1.7B                              & 1.72B  & 2048 & 68.52 & 71.63 & 65.21 & 78.76 \\
    codefuse-ai/F2LLM-v2-4B                                & 4.02B  & 2560 & 69.46 & 72.41 & 67.06 & 80.15 \\
    Qwen/Qwen3-Embedding-4B                                & 4.02B  & 2560 & 70.13 & 74.61 & 69.45 & 80.07 \\
    Qwen/Qwen3-Embedding-8B                                & 7.57B  & 4096 & 71.42 & \textbf{75.23} & \textbf{70.58} & 80.69 \\
    codefuse-ai/F2LLM-v2-8B                                & 7.57B  & 4096 & 70.57 & 72.86 & 68.09 & 80.16 \\
    codefuse-ai/F2LLM-v2-14B                               & 13.99B & 5120 & 70.90 & 73.08 & 68.74 & \textbf{80.75} \\
    \midrule
    \textbf{Ours: 3B-0826}                                 & 3B     & 2048 & 74.56 & 71.93 & 63.89 & 76.93 \\
    \textbf{Ours: 10B-A1.8B-0826}                          & 10B    & 1536 & \textbf{74.98} & 72.23 & 65.64 & 78.41 \\
    \bottomrule
  \end{tabular}
\end{table*}

\Cref{tab:baselines,tab:large-baselines} compare Giga-Embeddings with public results from the \href{https://huggingface.co/spaces/mteb/leaderboard}{MTEB Leaderboard}.
Differences in prompts and evaluation settings make these contextual rather than controlled same-run comparisons.
Giga-Embeddings-instruct-10B-A1.8B-0826 leads the Russian column, while Qwen3-Embedding-8B leads English and Multilingual and F2LLM-v2-14B leads Code.

\section{Conclusion}

We introduce Giga-Embeddings-10B-A1.8B, to the best of our knowledge the first general-purpose text embedding encoder built around a sparse Mixture-of-Experts architecture.
By adapting an MoE backbone into a bidirectional encoder, the model combines 10B parameters of resident capacity with approximately 1.8B active parameters per token, providing an efficient architecture for high-throughput embedding inference.
It achieves the highest throughput among the five evaluated serving configurations while also obtaining the best aggregate scores in the Giga-Embeddings family across English, Russian, Multilingual, and Code benchmarks and leading the external comparison on Russian MTEB.
These results show that encoder MoE models can improve inference throughput without sacrificing embedding quality.
At the compact end of the family, similarity-distribution distillation produces consistent gains of 0.09--0.22 points for the 480M model while remaining independent of teacher and student embedding dimensions.

\section{Limitations}

Our benchmark results report a single run per model, so small score differences do not have uncertainty estimates.
The throughput measurements are specific to one serving environment, omit the 480M model, and do not isolate the contribution of sparse activation from other architectural and implementation differences.
The training mixture contains non-public data, which limits full reproduction and independent auditing for benchmark contamination.
Finally, aggregate MTEB scores can hide variation across individual tasks and languages.
Future evaluations should include matched dense--sparse controls and report latency, peak memory, and energy use across a broader range of hardware and serving configurations.

\FloatBarrier
\bibliography{references}
\bibliographystyle{icml2026}

\end{document}